\documentclass[11pt]{article}

\usepackage[margin=1in]{geometry}
\usepackage{booktabs}
\usepackage{amsmath}
\usepackage{array}
\usepackage{xcolor}
\usepackage{microtype}
\usepackage{url}
\usepackage{graphicx}
\usepackage{float}
\usepackage{tikz}
\usetikzlibrary{arrows.meta, positioning}
\usepackage[hidelinks]{hyperref}
\newcommand{\dataset}{Curate-DTI}

\title{\textbf{A curated multi-endpoint drug–protein bioactivity dataset with construct-aware protein targets}}

\usepackage{authblk}

\author[1,*]{Lina Abou-Abbas}
\author[1]{Nour Mhanna}
\author[1]{Hasan Bazzi}
\author[1]{Sleiman El Hajj}

\author[1]{Abdulkader Helwan}
\author[2]{Rabah Omrani }
\author[2]{Khadidja Henni}

\affil[1]{Department of Electrical and Computer Engineering, Lebanese American University, Byblos, Lebanon}
\affil[2]{Institute of Applied Artificial Intelligence, T\'ELUQ University, Montreal, Canada}
\affil[*]{Corresponding author: \texttt{lina.abouabbas@lau.edu.lb}}

\date{}

\begin{document}
\maketitle

\begin{abstract}

Public drug–protein bioactivity resources commonly identify targets by UniProt accession, although one accession may represent distinct experimental sequences. We present Curate-DTI, a construct-aware dataset derived from the 1 July 2026 BindingDB export, with IC50, Ki, Kd and EC50 processed independently. The release contains 2,919,717 measurements, 1,291,872 standardized ligands, 9,668 sequence-defined protein entities and 2,275,683 endpoint-specific interactions spanning 8,336 reviewed UniProt accessions. Protein entities combine the accession with the reported experimental sequence, preserving constructs that accession-only processing would merge. Across the four endpoints, 4,204 endpoint-specific ligand–accession groups showed different classification labels across sequence-defined protein entities. Exact and inequality-qualified observations provide relation-aware classification evidence, whereas regression targets use exact values only. Connected Parquet tables preserve individual measurements, standardized structures, protein sequences, aggregated interactions and available assay and source provenance. The resource supports endpoint-specific classification and regression while allowing users to select molecular representations, protein representations and evaluation splits. Preprocessing code, a column dictionary, exclusion reports and validation summaries accompany the release.

\end{abstract}

\section{Background \& Summary}

BindingDB is a large public repository of experimentally measured bioactivities
between small molecules and protein targets
\cite{gilson2016bindingdb,liu2025bindingdb}. It has been used to develop and
evaluate computational models of drug--target interaction and quantitative
drug--protein activity
\cite{abou2026protein,rifaioglu2019recent,karimi2019deepaffinity,
khokhlov2025drugform,ouyang2025improving}. However, its downloadable export
requires substantial processing before consistent machine-learning use. It
contains several experimental endpoints, inequality-qualified measurements,
repeated observations, alternative chemical representations, incomplete assay
metadata and different descriptions of protein targets. Consequently, studies described as using BindingDB may retain
different records and produce different prediction targets depending on their
preprocessing decisions. \\
\\Although IC$_{50}$, \(K_i\), \(K_d\) and EC$_{50}$ are often grouped under the general term \emph{binding affinity}, they quantify different aspects of
drug--protein activity. \(K_d\) measures equilibrium binding, \(K_i\)
is the equilibrium inhibition constant inferred under a specified inhibition model, IC$_{50}$ measures the concentration required to reduce
a measured activity by half, and EC$_{50}$ measures the concentration required
to produce half of the maximum observed response \cite{buker2019enzyme,tonge2019quantifying}. Although these endpoints may be related under specific mechanistic assumptions
and experimental conditions, their reported values cannot generally be
converted from one endpoint to another using the numerical value alone. They
were therefore retained as separate endpoint-specific prediction targets. The present resource identifies each
endpoint explicitly and processes the four endpoints independently rather than
combining them into a common regression target.
\\
\\
A separate issue arises even after the endpoint types are kept distinct:
measurements of the same endpoint for the same compound and nominal target may
differ across assays. For example, two reported IC$_{50}$ values may disagree
with each other because they were obtained under different experimental
conditions; similarly, independently reported \(K_i\) values may differ.
Landrum and Riniker documented substantial disagreement when literature
IC$_{50}$ measurements were combined across sources and when literature
\(K_i\) measurements were combined across sources
\cite{landrum2024combining}. The resource therefore preserves each individual
measurement, its reported relation and its available assay and source
information. Repeated measurements are summarized separately within each
endpoint, and the interaction tables identify conflicting classification
evidence and widely dispersed exact values.
\\
\\Protein identity presents an additional curation challenge. A reviewed UniProt
accession identifies a curated Swiss-Prot protein record and its canonical
sequence \cite{uniprot2025uniprot}. However, measurements associated with one
accession may involve different experimental sequences, including the complete
protein, an isolated domain, a truncated construct, a substitution variant or
an engineered sequence. Aggregating targets only by accession can therefore
merge measurements obtained using distinct experimental proteins. In this
resource, a protein entity is defined using both its reviewed Swiss-Prot
accession and its normalized experimental sequence. Target names and organism
descriptions are retained as metadata but do not determine protein identity.
\\
The final release contains 469 reviewed UniProt accessions associated with more
than one retained sequence-defined protein entity. Across the four
endpoint-specific interaction views, 4,204 drug--accession groups have
classification outcomes that depend on the experimental protein sequence.
Collapsing these sequences to their shared accession can therefore merge
distinct experimental evidence and create additional classification conflicts.
Construct-aware protein identity preserves these sequence-level distinctions.
\\
\\Existing resources such as Davis and KIBA have enabled common comparisons of quantitative
drug--target prediction methods
\cite{davis2011comprehensive,tang2014making,thafar2022affinity2vec,
ozturk2018deepdta,shim2021prediction}. However, their targets are not
identical: Davis is based primarily on \(K_d\), whereas KIBA uses an integrated
bioactivity score. Broader benchmarking initiatives such as MoleculeNet have similarly
demonstrated the importance of documented preprocessing, evaluation metrics
and splitting procedures
\cite{wu2018moleculenet,li2026systematic}. 
\\
\\Large curated bioactivity resources provide complementary approaches to data
reuse. ChEMBL \cite{zdrazil2024chembl} combines manually curated activity, assay and target information
and includes curated annotations for a subset of protein variants, while
Papyrus \cite{bequignon2023papyrus} integrates and standardizes bioactivity records from several public
sources. Therapeutics Data Commons \cite{huang2021therapeutics} provides task-oriented datasets, processing
utilities and evaluation procedures. The present release instead focuses on a
defined BindingDB export, preserving individual relation-qualified
measurements and available source provenance while defining every retained
protein target using its reviewed accession and BindingDB-reported experimental
sequence.
\\A large BindingDB-derived resource
additionally requires explicit treatment of endpoint meaning, repeated
evidence, experimental protein constructs and source provenance.
Evaluation results also depend on how measurements are partitioned. Random
measurement-level splits can place repeated observations, related compounds,
different constructs of the same accession or measurements from the same
article or patent in training, validation and test sets. Cold-drug,
cold-protein-entity, cold-UniProt, temporal and
source-document-grouped splits address different generalization questions
\cite{abou2026protein}. 
Canonical SMILES are provided so that users can calculate molecular-scaffold
groups using their preferred definition.
The release preserves the identifiers and provenance
needed to construct these partitions and to state explicitly which entities
are excluded from training.\\
\\
\dataset{} was developed as a documented and machine-readable resource derived
from the 1 July 2026 BindingDB release. It provides:

\begin{enumerate}
    \item separate IC$_{50}$, \(K_i\), \(K_d\) and EC$_{50}$ measurement
    and interaction tables;
    \item relation-aware preservation of exact and inequality-qualified
    measurements;
    \item standardized ligand structures shared across endpoints;
    \item construct-aware protein entities defined from reviewed accessions
    and experimental sequences;
    \item individual measurements linked to available assay and
    source-document provenance; and
     \item endpoint-specific classification and regression records supporting
    reproducible benchmarking.
\end{enumerate}

The primary interaction tables contain curated ligand bioactivities for
targets reported as exactly one protein chain, with a reviewed Swiss-Prot
accession and a valid experimental sequence. Multi-chain measurements and their
available chain-specific information are retained in separate archival tables
but are not included in the current modeling tables. The release does not
prescribe molecular descriptors, protein encoders or prediction architectures,
allowing researchers to select representations appropriate to their modeling
objectives.

\section{Methods}

\begin{figure}[H]
    \centering
    \includegraphics[
        width=\textwidth,
        height=0.88\textheight,
        keepaspectratio
    ]{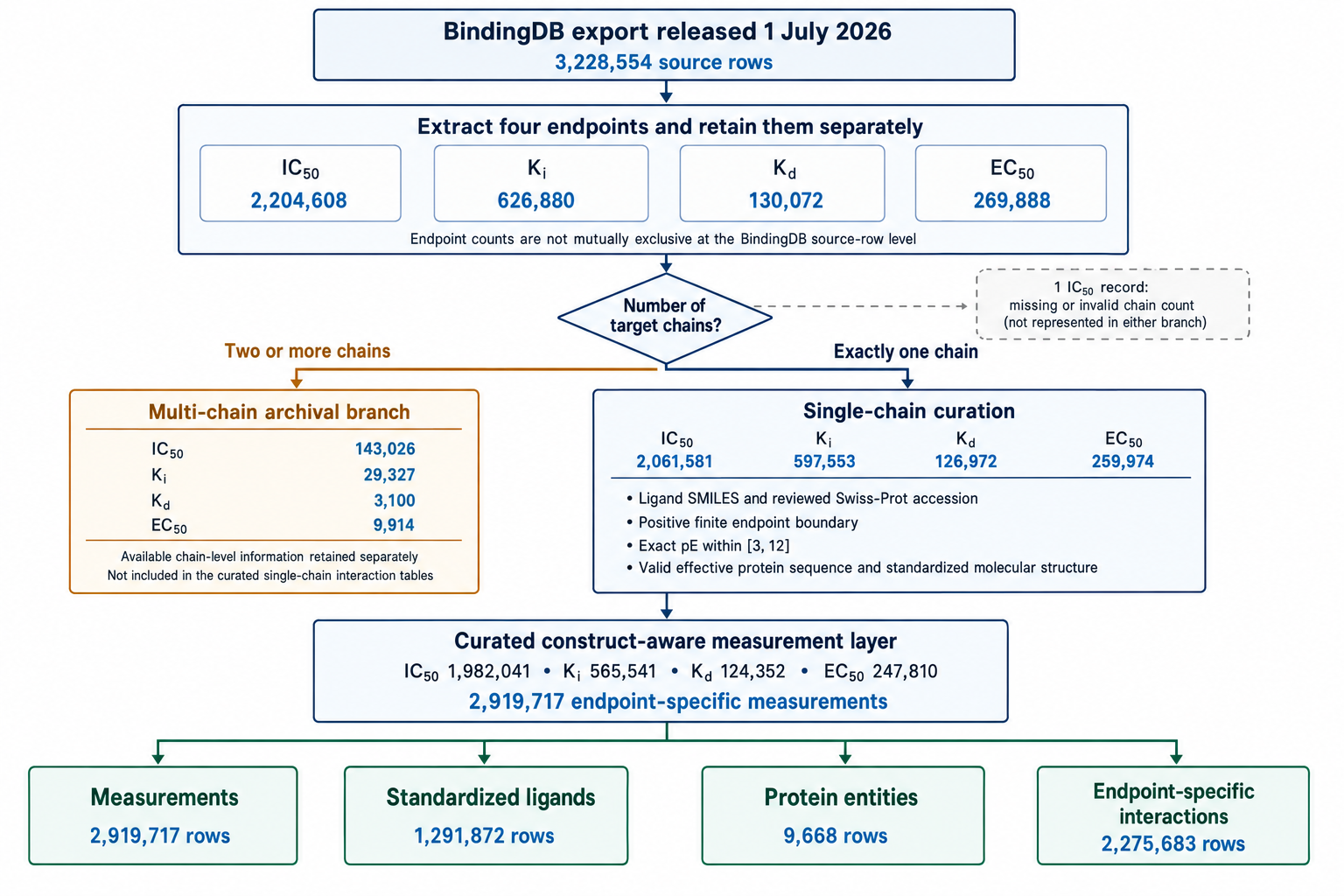}

    \caption{BindingDB preprocessing workflow for the four activity endpoints:
    IC$_{50}$, $K_i$, $K_d$, and EC$_{50}$. Measurements for each endpoint were
    processed independently. Single-chain records underwent activity-value
    validation, reviewed Swiss-Prot accession checking, experimental sequence
    validation, and ligand structure standardization. The resulting
    construct-aware measurements were then used to build shared ligand and
    protein-entity tables together with endpoint-specific interaction datasets.
    Multi-chain measurements were retained separately, along with the available
    chain-level information.}

    \label{fig:preprocessing_workflow}
\end{figure}

The preprocessing workflow transformed the 1 July 2026 BindingDB export into
endpoint-specific measurement and interaction tables for classification and
regression studies. IC$_{50}$, \(K_i\), \(K_d\) and EC$_{50}$ observations
were extracted and processed independently. Individual experimental evidence
and available provenance were preserved while ligand structures were
standardized, protein targets were distinguished using their experimental
sequences and repeated measurements were aggregated into construct-aware
interactions. The complete workflow is summarized in
Fig.~\ref{fig:preprocessing_workflow}.
%
\subsection{Input data and release control}

The primary input was the complete BindingDB tab-separated export released on
1 July 2026 \cite{gilson2016bindingdb,liu2025bindingdb}. The file,
\texttt{bindingdb\_july\_2026.tsv}, was obtained from the BindingDB archive
hosted by the University of California San Diego Library. It contained
3,228,554 source rows. BindingDB contains experimentally reported binding and
activity measurements rather than molecular-docking scores or computationally
predicted binding energies.

Each source row was assigned a release-specific row identifier before
processing. IC$_{50}$, \(K_i\), \(K_d\) and EC$_{50}$ fields were then
examined independently because one BindingDB row can contain more than one
endpoint. The raw export contained 2,204,608 rows reporting IC$_{50}$,
626,880 reporting \(K_i\), 130,072 reporting \(K_d\) and 269,888 reporting
EC$_{50}$. These endpoint counts are therefore not mutually exclusive.

Protein accessions were validated against the reviewed Swiss-Prot FASTA file
\texttt{uniprot\_sprot.fasta.gz}, obtained from the UniProt release repository
on 23 July 2026 \cite{uniprot2025uniprot}. The file contained 575,503 reviewed
protein entries. For each matched accession, Swiss-Prot provided the canonical
amino-acid sequence and associated protein metadata, including the reviewed
protein name, entry name, organism and taxonomy identifier.

The experimental target-chain sequence reported by BindingDB was used to
define each sequence-specific protein entity. The corresponding canonical
Swiss-Prot sequence was used for accession validation, protein annotation and
construct characterization. All retained measurements had a target-chain
sequence reported by BindingDB.

Both source files were treated as fixed, read-only inputs. Their filenames,
download dates, file sizes and SHA-256 checksums are recorded in the release
manifest. All reported statistics were calculated from these fixed inputs.
The preprocessing pipeline produced a versioned release containing the curated
Parquet tables, multi-chain archival records, exclusion reports, validation
summaries, a column dictionary and the release manifest.
\subsection{Computational environment}

Preprocessing was performed using Python 3.12.4. The principal software
dependencies were pandas 2.2.3 and NumPy 1.26.4 for tabular and numerical
processing, RDKit 2025.09.4 for molecular parsing and standardization,
Biopython 1.86 for protein-sequence alignment, and fastparquet 2025.12.0 for Parquet serialization. The Parquet files were written using Zstandard compression.

The pipeline did not use random sampling, model training or stochastic data
transformations. Distinct input molecular structures were standardized in
batches, and intermediate molecular-standardization and protein-sequence
alignment results were cached. Cached results were indexed by their
corresponding inputs and processing parameters, allowing interrupted executions
to resume without altering the resulting curated records.

The deposited code includes a version-pinned environment specification, the
preprocessing pipeline, the scripts used to generate dataset statistics and
validation reports, and instructions for reconstructing the released files
from the fixed BindingDB and Swiss-Prot inputs.

\subsection{Measurement selection}

The four endpoint fields were processed independently. The raw BindingDB
export contained 2,204,608 IC$_{50}$ observations, 626,880 \(K_i\)
observations, 130,072 \(K_d\) observations and 269,888 EC$_{50}$
observations. These counts are not mutually exclusive because one source row
may report more than one endpoint. The endpoints were retained separately and
were not combined into a common regression target because they represent
different experimental quantities.

The model-ready tables were restricted to measurements involving targets
reported as exactly one protein chain. This criterion retained 2,061,581
IC$_{50}$, 597,553 \(K_i\), 126,972 \(K_d\) and 259,974 EC$_{50}$
observations. Multi-chain observations were excluded from the model-ready
tables because the BindingDB export did not provide sequences for all
constituent chains. They were retained separately in the archival tables,
together with their available chain-level information.

A single-chain observation was required to contain a ligand SMILES, a
reviewed Swiss-Prot primary accession and a positive, finite endpoint value or
inequality boundary. The accession was required to occur in the frozen
Swiss-Prot reference, and a valid experimental amino-acid sequence reported by BindingDB had to be
available. For exact measurements, the endpoint-specific transformed value defined in
Eq.~\ref{eq:pe_transformation} was required to lie within \([3,12]\). corresponding to an exact numerical
value between \(10^{-3}\) and \(10^{6}\) nM. This operational range check was
not applied to inequality-qualified measurements because a boundary outside
the interval may still provide logically valid classification evidence.

Ligand structures were subsequently standardized as described below.
Measurements were excluded when their reported ligand structure could not be
parsed or converted into a valid standardized parent structure.

The numbers retained after each sequential processing stage are presented in
Table~\ref{tab:attrition}.

\begin{table}[ht]
\centering
\caption{Endpoint-specific attrition during construction of the curated
single-chain measurement tables. Values indicate the number of measurements
retained after each sequential processing stage.}
\label{tab:attrition}
\small
{
\begin{tabular}{lrrrr}
\toprule
\textbf{Processing stage} &
\textbf{IC$_{50}$} &
\textbf{\(K_i\)} &
\textbf{\(K_d\)} &
\textbf{EC$_{50}$} \\
\midrule
Endpoint reported
& 2,204,608 & 626,880 & 130,072 & 269,888 \\
Exactly one target chain
& 2,061,581 & 597,553 & 126,972 & 259,974 \\
SMILES and primary accession reported
& 2,008,206 & 584,594 & 125,658 & 251,885 \\
Positive finite value or boundary
& 2,008,054 & 584,565 & 125,625 & 251,850 \\
Exact \(pE\) restricted to \([3,12]\)
& 2,002,416 & 580,881 & 124,729 & 251,246 \\
Accession present in reviewed Swiss-Prot
& 1,999,681 & 580,431 & 124,699 & 250,562 \\
Valid experimental amino-acid sequence
& 1,999,551 & 580,431 & 124,699 & 250,562 \\
Valid standardized ligand structure
& 1,982,041 & 565,541 & 124,325 & 247,810 \\
\bottomrule
\end{tabular}
}
\end{table}
\subsection{Relation-aware endpoint parsing and task labels}

Each retained IC$_{50}$, \(K_i\), \(K_d\) or EC$_{50}$ observation was
parsed independently into an endpoint type, a normalized relation
\(\{=,<,>\}\) and a positive numerical boundary expressed in nM. The original
reported value and BindingDB source-row identifier were retained so that the
parsing could be audited. Entries that could not be converted into a positive,
finite numerical boundary were excluded.

Exact observations were represented by \(=\), observations reporting a lower
boundary by \(>\), and observations reporting an upper boundary by \(<\).
The final relation counts are presented in
Table~\ref{tab:measurement_relations}.

\begin{table}[ht]
\centering
\caption{Relations among the retained endpoint-specific measurements.
Right-censored measurements report a lower boundary, whereas left-censored
measurements report an upper boundary.}
\label{tab:measurement_relations}
\small
\begin{tabular}{lrrrr}
\toprule
\textbf{Endpoint} &
\textbf{Measurements} &
\textbf{Exact (\(=\))} &
\textbf{Right-censored (\(>\))} &
\textbf{Left-censored (\(<\))} \\
\midrule
IC$_{50}$ & 1,982,041 & 1,561,575 & 254,428 & 166,038 \\
\(K_i\)   &   565,541 &   489,745 &  57,603 &  18,193 \\
\(K_d\)   &   124,325 &    77,589 &  39,357 &   7,379 \\
EC$_{50}$ &   247,810 &   196,715 &  40,772 &  10,323 \\
\midrule
Total     & 2,919,717 & 2,325,624 & 392,160 & 201,933 \\
\bottomrule
\end{tabular}
\end{table}

Exact measurements were eligible to provide continuous regression targets.
For endpoint \(E\), the transformed value was calculated as

\begin{equation}
pE = 9-\log_{10}(E_{\mathrm{nM}}).
\label{eq:pe_transformation}
\end{equation}

The corresponding regression targets are denoted pIC$_{50}$, \(pK_i\),
\(pK_d\) and pEC$_{50}$. Inequality-qualified measurements were retained as
experimental evidence but were not treated as exact regression observations.

For classification, a common operational threshold of 1,000 nM was applied
independently to each endpoint. An exact value less than or equal to
1,000 nM provided threshold-positive evidence (class 1), whereas an exact
value greater than 1,000 nM provided threshold-negative evidence (class 0).
An inequality-qualified observation was assigned a class only when its
boundary placed every possible value on one side of the threshold.
Exact measurements were retained in the standard regression view when
\(pE\in[3,12]\), corresponding to concentrations from 1 mM to 1 pM. This
broad interval covers the range normally relevant to quantitative
small-molecule bioactivity modeling while excluding exceptionally weak or
strong values that warrant additional scrutiny for possible unit,
transcription or parsing errors. The limits are operational quality-control
thresholds rather than universal biological or physical boundaries; valid
measurements may occur outside them in particular experimental settings.

The range restriction was applied only to exact measurements. It was not
applied to inequality-qualified measurements because an inequality outside
this interval may still provide valid classification evidence.
\begin{table}[ht]
\centering
\caption{Measurement-level classification rules at the operational threshold
of 1,000 nM.}
\label{tab:classification_evidence}
\small
\begin{tabular}{lll}
\toprule
\textbf{Relation} &
\textbf{Boundary} &
\textbf{Classification evidence} \\
\midrule
\(=\) & \(\leq 1{,}000\) nM & Class 1 \\
\(=\) & \(> 1{,}000\) nM & Class 0 \\
\(<\) & \(\leq 1{,}000\) nM & Class 1 \\
\(<\) & \(> 1{,}000\) nM & Ambiguous \\
\(>\) & \(\geq 1{,}000\) nM & Class 0 \\
\(>\) & \(< 1{,}000\) nM & Ambiguous \\
\bottomrule
\end{tabular}
\end{table}

For example, \(E<500\) nM provides class-1 evidence and
\(E>5{,}000\) nM provides class-0 evidence. In contrast,
\(E<5{,}000\) nM and \(E>500\) nM are ambiguous because the unknown exact
value could lie on either side of the threshold. Ambiguous observations were
retained but were not assigned a binary measurement-level label.

The same numerical threshold was used to provide consistent task definitions
across the four endpoint-specific datasets. This does not imply that
IC$_{50}$, \(K_i\), \(K_d\) and EC$_{50}$ have equivalent biological
interpretations. Class 0 indicates only that the reported endpoint evidence
lies above the operational threshold; it should not be interpreted as proof
that the ligand is biologically inactive under all experimental conditions.

\subsection{Ligand-structure standardization}

Ligand structures were standardized using RDKit
\cite{landrum2025rdkit}. Molecular cleanup, parent-fragment selection, charge
neutralization and sanitization were applied before generating canonical
isomeric SMILES, standardized InChI and InChIKey representations. Defined
stereochemistry was preserved, whereas unspecified stereochemistry was not
inferred.

The canonical isomeric SMILES defined ligand identity. Structures producing
the same standardized SMILES were assigned the same \texttt{Drug\_ID}, which
was generated from a SHA-256 hash of the standardized SMILES. Molecular
weight, heavy-atom count, carbon presence, atom count, bond count and formal
charge were retained as descriptive fields but were not used as inclusion
criteria. Therefore, no molecular-weight, heavy-atom-count or carbon-content
restriction was applied.

Structures were excluded only when they could not be parsed or standardized
into a valid parent structure. Excluded structures and their corresponding
failure reasons are provided in the exclusion report. Available ligand names
and external database identifiers were retained for cross-referencing but did
not determine chemical identity.

\subsection{Construct-aware protein entities}

Protein identity was defined using both the reviewed UniProt accession and the
experimental target-chain sequence reported by BindingDB. Sequences were
normalized by removing whitespace and converting letters to uppercase.
Measurements sharing the same accession and normalized sequence were assigned
to the same protein entity, whereas different sequences associated with the
same accession remained distinct.

Each \texttt{Protein\_Entity\_ID} consisted of the UniProt accession, two
underscore characters and the first 16 hexadecimal characters of the SHA-256
hash of the normalized sequence:

\begin{equation}
\texttt{Protein\_Entity\_ID}
=
\texttt{UniProt\_ID}
\mathbin{\Vert}
\texttt{\char`\_\char`\_}
\mathbin{\Vert}
\texttt{SHA256(Sequence)[0:16]},
\end{equation}
where \(\Vert\) denotes string concatenation and
\texttt{\char`\_\char`\_} is the literal two-underscore separator.

The complete sequence hash was also retained. Protein identity therefore did
not depend on potentially inconsistent target names or organism descriptions.

Each experimental sequence was compared with the canonical sequence of its
corresponding Swiss-Prot entry. Exact canonical matches, contiguous fragments
and terminal extensions were identified directly. Other sequences were
compared using Biopython \texttt{PairwiseAligner} with the BLOSUM62
substitution matrix, a gap-opening score of \(-11\) and a gap-extension score
of \(-1\) \cite{cock2009biopython}. Sequences whose length was at least 90\% of the canonical sequence length were
aligned globally, whereas shorter sequences were aligned with unpenalized
terminal gaps. The resulting annotations identify fragments, substitution variants,
insertion/deletion variants and terminal extensions. These annotations are
descriptive and do not determine protein identity. Complete category
definitions and assignment order are provided in the preprocessing code and
release manifest.

The final release contains 9,668 sequence-defined protein entities associated
with 8,336 reviewed UniProt accessions. All 2,919,717 retained measurements
were associated with an experimental target sequence reported by BindingDB.
Swiss-Prot sequences were used only as canonical references for accession
validation, protein annotation and sequence comparison.

\subsection{Interaction aggregation and task views}

Measurements were aggregated independently for IC$_{50}$, \(K_i\), \(K_d\)
and EC$_{50}\). Within each endpoint, one interaction record was created for
each unique combination of \texttt{Drug\_ID} and
\texttt{Protein\_Entity\_ID}. Measurements from different endpoints were not
combined. Each interaction reports the numbers of contributing exact,
left-censored and right-censored measurements.

For classification, measurement-level evidence was combined at the interaction
level. A valid class-1 label was assigned when class-1 evidence was present
without class-0 evidence, and a valid class-0 label was assigned when class-0
evidence was present without class-1 evidence. Interactions containing evidence
for both classes were marked as conflicts. Interactions supported only by
ambiguous measurements were retained without a valid classification label.
Users should select records with \texttt{classification\_valid = true} and use
\texttt{y\_class} as the classification target.

Regression summaries were calculated only from exact measurements. The median
exact value in nM was calculated for each interaction and transformed into the
corresponding pIC$_{50}$, \(pK_i\), \(pK_d\) or pEC$_{50}$ target. Interactions
without an exact measurement were retained but were not assigned a valid
regression target. Users should select records with
\texttt{regression\_valid = true} for regression.

Interactions whose exact measurements spanned at least 100-fold were flagged
as dispersed but were not excluded. A deterministic
\texttt{Interaction\_ID} was generated from the endpoint type,
\texttt{Drug\_ID} and \texttt{Protein\_Entity\_ID}.
\\All exact-value summary statistics, including the median, minimum, maximum and
interquartile range, were calculated in nM. The median in nM was subsequently
transformed to the endpoint-specific \(pE\) regression target. The dispersed
flag identifies interactions whose exact values span at least 100-fold,
equivalent to a difference of at least two \(pE\) units.
\subsection{Measurement provenance and source grouping}

Each retained measurement preserves its BindingDB source-row identifier,
available assay information and bibliographic provenance, including article
DOI, PMID, patent number and BindingDB entry DOI. A
\texttt{Source\_Document\_ID} was generated to group measurements originating
from the same source document.

Source documents were identified using normalized article DOI when available,
followed by PMID, patent number and BindingDB entry DOI. When none of these
identifiers was available, a release-specific fallback identifier was assigned
so that unrelated measurements with missing provenance were not placed in one
shared group.

Source-document identifiers are retained in the measurement tables to support
auditing and source-grouped data splitting. Measurements from the same article
or patent can therefore be assigned to the same training, validation or test
partition, reducing information leakage across evaluation sets.

\subsection{Release organization and validation}

The processed release was organized into shared, endpoint-specific, archival
and reporting directories. The shared directory contains the standardized
ligand table, construct-aware protein-entity table and combined long-form
measurement and interaction tables. Separate endpoint directories provide
IC$_{50}$, \(K_i\), \(K_d\) and EC$_{50}$ measurement and interaction views.
Multi-chain measurements and their available chain-level information are
stored separately as archival data.

All primary tables were written in Parquet format using Zstandard compression.
A column dictionary describes the meaning and data type of each released
field. The release also includes a manifest recording the source files,
software versions, processing parameters, endpoint definitions and generated
files. Exclusion and statistical reports document records removed or retained
at each processing stage.

Automated validation was performed after preprocessing. The checks confirmed
that measurement, drug, protein-entity and interaction identifiers were
unique within their corresponding tables. Every drug and protein entity
referenced by a measurement or interaction record was present in the shared
drug and protein tables. Each interaction was unique for its endpoint,
\texttt{Drug\_ID} and \texttt{Protein\_Entity\_ID}. Valid classification
records contained a binary label, and valid regression records contained a
finite endpoint-specific target. The released row counts were also compared
with the preprocessing and endpoint summary reports. All final validation
checks passed.

\section{Data Overview}

\subsection{Dataset composition}

The curated release contains 2,919,717 individual bioactivity measurements:
1,982,041 IC$_{50}$, 565,541 \(K_i\), 124,325 \(K_d\) and 247,810
EC$_{50}$ observations. Each measurement is retained separately with its
reported relation, endpoint value and available assay and source provenance.

Measurements involving the same standardized ligand, sequence-defined protein
entity and endpoint were aggregated into interaction records. This produced
2,275,683 endpoint-specific interactions, comprising 1,548,382 IC$_{50}$,
445,298 \(K_i\), 86,858 \(K_d\) and 195,145 EC$_{50}$ interactions. Across
the four endpoints, these records represent 2,193,251 distinct construct-aware
drug--protein pairs. The measurement tables preserve individual experimental
observations, whereas the interaction tables provide classification and
regression targets derived from repeated evidence.

The release contains 1,291,872 standardized ligands and 9,668 sequence-defined
protein entities associated with 8,336 reviewed UniProt accessions. The number
of protein entities exceeds the number of accessions because measurements
assigned to one accession may involve different experimental sequences,
including canonical proteins, fragments, variants and engineered constructs.
A total of 469 accessions are associated with more than one retained protein
entity.

Across the four endpoint-specific datasets, construct-aware aggregation
produced 2,275,683 interaction records, compared with 2,212,342 records when
protein targets were collapsed to UniProt accessions. Construct-aware identity
therefore preserved 63,341 endpoint-specific interactions that would otherwise
have been merged. A total of 4,204 endpoint-specific drug--accession groups
had classification outcomes that depended on the experimental protein
sequence.

The protein entities represent 431 organism labels. Human proteins constitute
the largest group, with 6,050 entities. Proteins from other organisms were
retained to preserve the broader experimental scope of BindingDB. Organism and
taxonomy annotations were obtained from the reviewed Swiss-Prot reference. The principal dataset counts are summarized in Table~\ref{tab:overview}.

\begin{table}[ht]
\centering
\caption{Overview of the multi-endpoint construct-aware dataset.}
\label{tab:overview}
\small
\begin{tabular}{lr}
\toprule
\textbf{Quantity} & \textbf{Count} \\
\midrule
Retained measurements & 2,919,717 \\
Endpoint-specific interactions & 2,275,683 \\
Distinct drug--protein pairs across endpoints & 2,193,251 \\
Standardized ligands & 1,291,872 \\
Sequence-defined protein entities & 9,668 \\
Reviewed UniProt accessions & 8,336 \\
Accessions with multiple protein entities & 469 \\
Organism labels & 431 \\
Valid classification interactions & 2,205,683 \\
Class-1 interactions & 1,362,117 \\
Class-0 interactions & 843,566 \\
Classification-conflict interactions & 38,687 \\
Valid regression interactions & 1,814,083 \\
\bottomrule
\end{tabular}
\end{table}

\subsection{Organism coverage}

The 9,668 sequence-defined protein entities are associated with 431 organism
labels obtained from the reviewed Swiss-Prot reference. No organism-based
restriction was applied during preprocessing. The released protein table
retains the organism name and taxonomy identifier, allowing users to construct
human-only, organism-specific or mixed-organism datasets.

Table~\ref{tab:organism_coverage} presents organism coverage at the protein
level. Human proteins form the largest group, with 6,050 protein entities,
followed by mouse and rat with 869 and 841 entities, respectively.

\begin{figure}[H]
\centering
\includegraphics[
    width=\textwidth,
    keepaspectratio
]{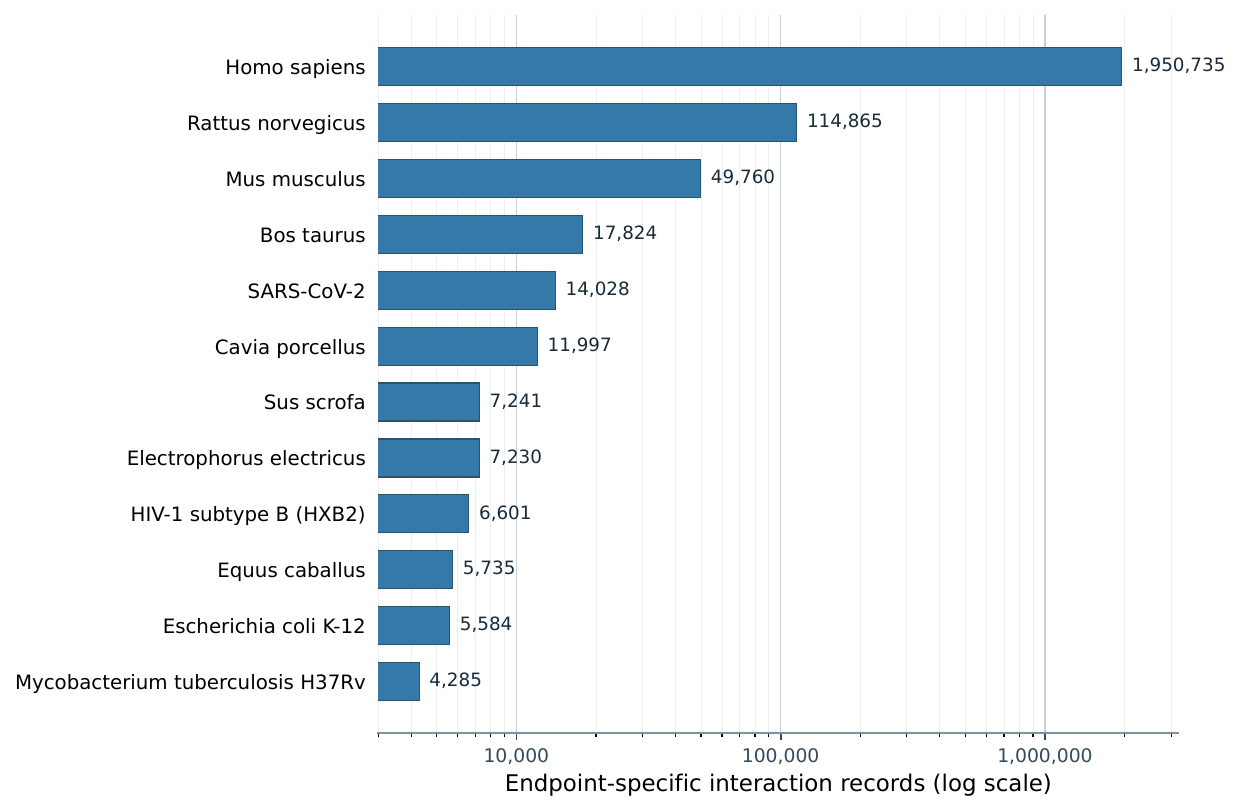}
\caption{Organism coverage of the endpoint-specific interaction records.
The twelve most represented organism labels are shown. Counts combine the
IC$_{50}$, \(K_i\), \(K_d\) and EC$_{50}$ interaction tables; therefore, a
drug--protein pair reported for more than one endpoint contributes one
interaction record to each corresponding endpoint. }
\label{fig:organism_interaction_coverage}
\end{figure}

Figure~\ref{fig:organism_interaction_coverage} presents coverage at the
interaction level. Human proteins are involved in 1,950,735 of the 2,275,683
endpoint-specific interaction records (85.72\%). Rat and mouse proteins
account for 114,865 and 49,760 interaction records, respectively. Protein and
interaction counts provide complementary information: the protein counts
describe target diversity, whereas the interaction counts reflect the amount
of bioactivity evidence available for modeling each organism.

\begin{table}[ht]
\centering
\caption{Most represented organism labels among the sequence-defined protein
entities. Counts describe protein entities rather than measurements or
interactions.}
\label{tab:organism_coverage}
\small
\begin{tabular}{lrr}
\toprule
\textbf{Organism} &
\textbf{Protein entities} &
\textbf{Percentage (\%)} \\
\midrule
\textit{Homo sapiens} & 6,050 & 62.58 \\
\textit{Mus musculus} & 869 & 8.99 \\
\textit{Rattus norvegicus} & 841 & 8.70 \\
\textit{Bos taurus} & 201 & 2.08 \\
\textit{Escherichia coli} strain K12 & 137 & 1.42 \\
\textit{Mycobacterium tuberculosis} H37Rv & 100 & 1.03 \\
\textit{Sus scrofa} & 85 & 0.88 \\
\textit{Saccharomyces cerevisiae} S288c & 76 & 0.79 \\
\textit{Oryctolagus cuniculus} & 58 & 0.60 \\
\textit{Cavia porcellus} & 43 & 0.44 \\
Other organism labels & 1,208 & 12.49 \\
\midrule
Total & 9,668 & 100.00 \\
\bottomrule
\end{tabular}
\end{table}

\subsection{Source provenance}

Available source information was preserved for every retained measurement.
Measurements were grouped using article DOI, PMID, patent number or BindingDB
entry DOI, according to the hierarchy described in the Methods. When no
document-level identifier was available, a release-specific fallback group was
assigned.

The 2,919,717 retained measurements are associated with 53,046 source groups.
Article DOI provides the grouping identifier for 1,420,613 measurements
(48.66\%), while patents account for 1,228,477 measurements (42.08\%). Only
705 measurements (0.02\%) required a fallback identifier because no
document-level identifier was available
(Table~\ref{tab:source_provenance}).

\begin{table}[ht]
\centering
\caption{Source provenance of the retained measurement records.
\\Percentages are rounded independently and may not sum exactly to 100\%.}
\label{tab:source_provenance}
\small
\begin{tabular}{lrrr}
\toprule
\textbf{Source-group type} &
\textbf{Measurements} &
\textbf{Unique groups} &
\textbf{Measurements (\%)} \\
\midrule
Article DOI & 1,420,613 & 38,263 & 48.66 \\
Patent & 1,228,477 & 8,218 & 42.08 \\
PMID & 169,782 & 4,787 & 5.82 \\
BindingDB entry DOI & 100,072 & 1,072 & 3.43 \\
PMID associated with multiple DOI values & 68 & 1 & $<0.01$ \\
Release-specific fallback & 705 & 705 & 0.02 \\
\midrule
Total & 2,919,717 & 53,046 & 100.00 \\
\bottomrule

\end{tabular}
\end{table}
\section{Data Records}

\subsection{Released data}

\begin{figure}[ht]
\centering
\includegraphics[
    width=\textwidth,
    keepaspectratio
]{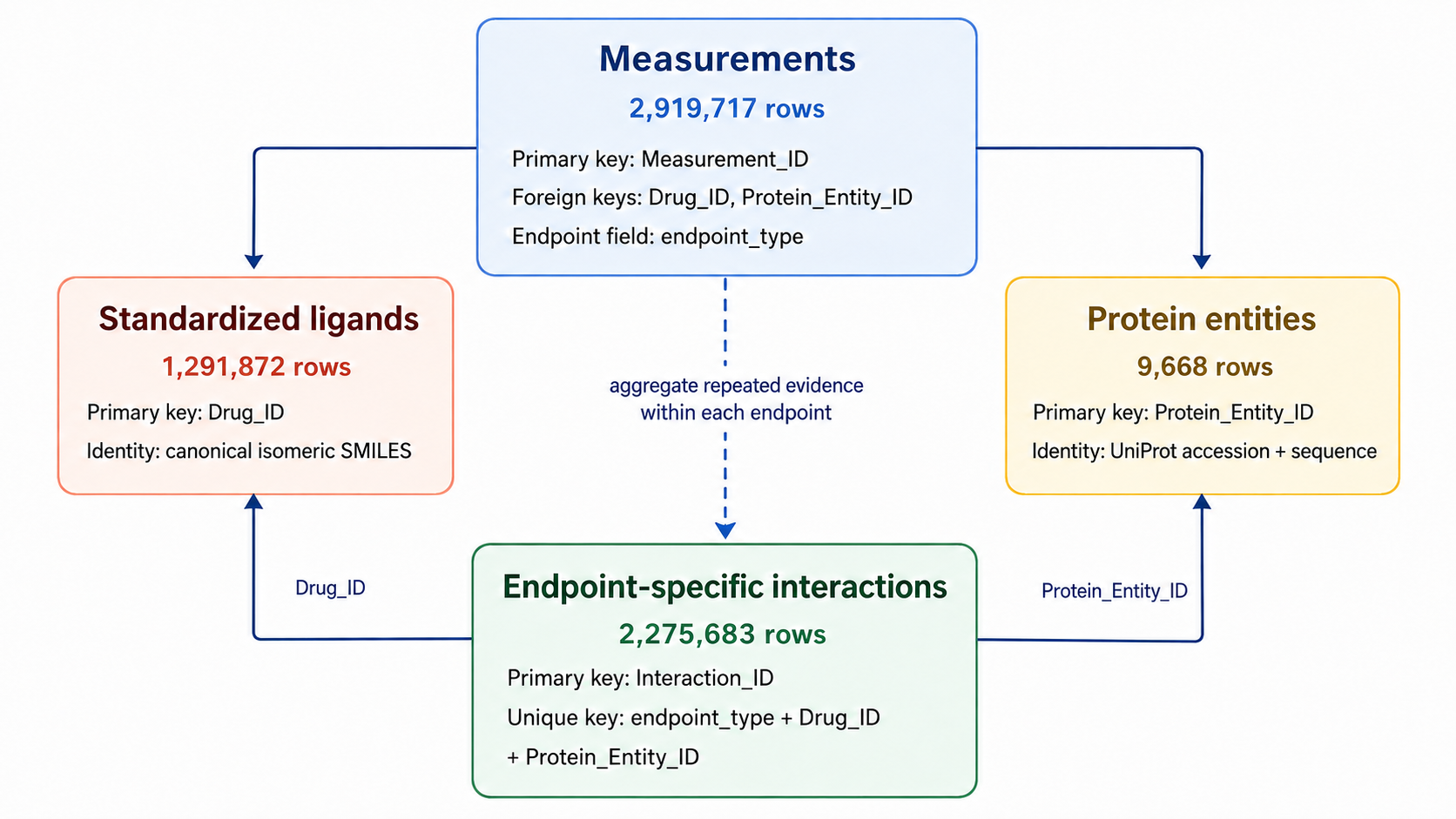}
\caption{Relationships among the four principal record types in the
model-ready release. Measurement and interaction records reference the shared
ligand and protein-entity tables through \texttt{Drug\_ID} and
\texttt{Protein\_Entity\_ID}. Repeated measurements are aggregated separately
within each endpoint to produce endpoint-specific interaction records.}
\label{fig:data_organization}
\end{figure}

The model-ready release is organized around four connected record types:
individual measurements, standardized ligands, sequence-defined protein
entities and endpoint-specific interactions
(Fig.~\ref{fig:data_organization}). The combined tables contain 2,919,717
measurements, 1,291,872 standardized ligands, 9,668 protein entities and
2,275,683 endpoint-specific interactions.

Each measurement is identified by \texttt{Measurement\_ID} and linked to one
ligand and one protein entity through \texttt{Drug\_ID} and
\texttt{Protein\_Entity\_ID}. Repeated measurements sharing the same endpoint,
ligand and protein entity are summarized in one interaction record identified
by \texttt{Interaction\_ID}. Separate IC$_{50}$, \(K_i\), \(K_d\) and
EC$_{50}$ measurement and interaction files are provided alongside the
combined multi-endpoint tables.
Table~\ref{tab:released_tables} summarizes the primary and supplementary data tables.

\begin{table}[H]
\centering
\caption{Primary model-ready tables included in the data release.}
\label{tab:released_tables}
\scriptsize
\begin{tabular}{p{0.19\linewidth}p{0.12\linewidth}
                p{0.22\linewidth}p{0.36\linewidth}}
\toprule
\textbf{Data table} &
\textbf{Rows} &
\textbf{Primary key} &
\textbf{Purpose} \\
\midrule

Measurements &
2,919,717 &
\texttt{Measurement\_ID}
&
Individual retained single-chain IC$_{50}$, \(K_i\), \(K_d\) and EC$_{50}$
observations, including reported relations, task evidence and available assay
and source provenance. \\

Standardized ligands &
1,291,872 &
\texttt{Drug\_ID}
&
One record for each standardized ligand structure, including canonical
chemical representations, calculated properties and available external
identifiers. \\

Protein entities &
9,668 &
\texttt{Protein\_Entity\_ID}
&
One record for each sequence-defined protein entity, including experimental
and canonical sequences, reviewed Swiss-Prot annotations and construct
characterization. \\

Endpoint-specific interactions &
2,275,683 &
\texttt{Interaction\_ID}
&
One aggregated record for each endpoint, ligand and protein-entity combination,
including classification and regression task fields and quality indicators. \\

\bottomrule
\end{tabular}
\end{table}

Supplementary archival tables preserve records outside the model-ready
single-chain dataset, including all endpoint-explicit observations and the
available chain-level information for multi-chain targets. These files are
provided for provenance and possible future curation and are described
separately below.
\subsection{Measurement records}

The combined measurement file,
\texttt{BindingDB\_single\_chain\_measurements\_all\_endpoints.parquet},
contains 2,919,717 rows and 48 columns. Each row represents one retained
experimental observation and is uniquely identified by
\texttt{Measurement\_ID}. The fields \texttt{Drug\_ID} and
\texttt{Protein\_Entity\_ID} link the measurement to the shared ligand and
protein-entity tables.

Each record identifies the endpoint type, original reported value, normalized
relation, numerical boundary in nM and transformed \(pE\) boundary.
Measurement-level fields indicate whether the observation is exact and whether
it provides class-1, class-0 or ambiguous evidence at the 1,000 nM threshold.
Inequality-qualified measurements have no exact regression target.

Available experimental and source information is retained, including assay pH
and temperature, BindingDB source-row and reactant-set identifiers, article
DOI, PMID, patent number, BindingDB entry DOI, publication date and
\texttt{Source\_Document\_ID}. Missing values indicate that the corresponding
information was not available in the BindingDB export.

Separate endpoint-specific files are provided for IC$_{50}$, \(K_i\), \(K_d\)
and EC$_{50}$. These files contain the same curated measurements with
endpoint-specific field names for direct use. The measurement records allow
users to inspect repeated evidence, assay variability, inequalities and source
provenance; the interaction tables described below provide the aggregated
classification and regression targets.

\subsection{Ligand records}

The shared ligand file,
\texttt{common/BindingDB\_drug\_structures.parquet}, contains 1,291,872 rows
and 32 columns. Each row represents one standardized ligand structure and is
uniquely identified by \texttt{Drug\_ID}, which links the table to the
measurement and interaction records.

The table provides canonical isomeric SMILES, standardized InChI and InChIKey,
the complete structure hash and calculated molecular properties, including
molecular weight, heavy-atom count, atom count, bond count, formal charge and
the presence of carbon. These properties are descriptive and were not used to
restrict inclusion in the final dataset.

Available BindingDB ligand names, monomer identifiers and external identifiers
from PubChem, ChEBI, ChEMBL, DrugBank, IUPHAR, KEGG, ZINC and the Protein Data
Bank are retained for interpretation and cross-referencing. Missing values
indicate that the corresponding identifier was not available in the BindingDB
export. Chemical identity within the release is determined by the standardized
canonical isomeric SMILES rather than by ligand names or external identifiers.

\subsection{Protein-entity records}

The shared protein file,
\texttt{common/BindingDB\_protein\_entities.parquet}, contains 9,668 rows and
48 columns. Each row represents one sequence-defined protein entity and is
uniquely identified by \texttt{Protein\_Entity\_ID}, which links the table to
the measurement and interaction records.

The table provides the reviewed UniProt accession, the experimental
target-chain sequence reported by BindingDB and the corresponding canonical
Swiss-Prot sequence. It also includes the reviewed protein name, entry name,
organism, taxonomy identifier, sequence lengths and complete sequence hashes.
All retained protein entities were defined using BindingDB-reported target
sequences.

Construct-characterization fields describe the relationship between the
experimental and canonical sequences. These include sequence category,
identity, canonical and experimental coverage, alignment coordinates and the
numbers of substitutions, insertions and deletions. Flags identify fragments,
terminal extensions and possible polyhistidine extensions. These fields are
descriptive and do not determine protein identity.

Available BindingDB target names, organism descriptions, Swiss-Prot names,
secondary accessions and target-chain PDB identifiers are retained as metadata.
Protein identity is determined only by the reviewed UniProt accession and
normalized experimental sequence.

\subsection{Construct-aware interaction records}

The combined interaction file,
\texttt{BindingDB\_single\_chain\_interactions\_all\_endpoints.parquet},
contains 2,275,683 rows and 54 columns. Each row summarizes measurements
sharing the same endpoint, \texttt{Drug\_ID} and
\texttt{Protein\_Entity\_ID}. Every row is uniquely identified by
\texttt{Interaction\_ID}.

The table reports the total number of contributing measurements and the numbers
of exact, left-censored and right-censored observations. Classification fields
include the counts of class-1, class-0 and ambiguous evidence,
\texttt{classification\_conflict}, \texttt{classification\_valid} and the
binary target \texttt{y\_class}. Interactions with conflicting evidence or
only ambiguous measurements are retained but have no valid classification
target.

Regression fields are calculated only from exact measurements. They include
the number, median, minimum, maximum and interquartile range of exact endpoint
values, the endpoint-specific transformed target and indicators of regression
validity and measurement dispersion. Interactions without an exact
measurement remain available but have no valid regression target.
The standard regression view excludes inequality-qualified measurements.
However, the retained relation and boundary fields allow users to develop
censored-regression or interval-aware models when appropriate.
Separate interaction files are provided for IC$_{50}$, \(K_i\), \(K_d\) and
EC$_{50}\). For classification, users should select records with
\texttt{classification\_valid = true} and use \texttt{y\_class}. For
regression, users should select records with
\texttt{regression\_valid = true} and use the corresponding pIC$_{50}$,
\(pK_i\), \(pK_d\) or pEC$_{50}$ target. Source-count fields indicate how many
measurements and source groups support each interaction; detailed provenance
remains available in the measurement tables.

\begin{table}[ht]
\centering
\caption{Endpoint-specific interaction files and prediction targets.}
\label{tab:modeling_files}
\small
\setlength{\tabcolsep}{4pt}
\begin{tabular}{p{0.12\linewidth}p{0.40\linewidth}
                p{0.21\linewidth}p{0.17\linewidth}}
\toprule
\textbf{Endpoint} &
\textbf{Interaction file} &
\textbf{Classification target} &
\shortstack{\textbf{Regression}\\\textbf{target}} \\
\midrule

IC$_{50}$ &
\shortstack[l]{\texttt{BindingDB\_ic50\_}\\
\texttt{interactions.parquet}} &
\texttt{y\_class} &
\texttt{pIC50} \\

\(K_i\) &
\shortstack[l]{\texttt{BindingDB\_ki\_}\\
\texttt{interactions.parquet}} &
\texttt{y\_class} &
\texttt{pKi} \\

\(K_d\) &
\shortstack[l]{\texttt{BindingDB\_kd\_}\\
\texttt{interactions.parquet}} &
\texttt{y\_class} &
\texttt{pKd} \\

EC$_{50}$ &
\shortstack[l]{\texttt{BindingDB\_ec50\_}\\
\texttt{interactions.parquet}} &
\texttt{y\_class} &
\texttt{pEC50} \\

\bottomrule
\end{tabular}
\end{table}

For classification, rows should be selected using
\texttt{classification\_valid = true}. For regression, rows should be selected
using \texttt{regression\_valid = true}. The four endpoints should be modeled
separately because their numerical targets have different experimental
interpretations.
\subsection{Archival records and supporting files}

The file
\texttt{data/BindingDB\_measurements\_all\_endpoints.parquet} contains
3,231,448 endpoint-explicit observations before application of the
single-chain eligibility and quality criteria. This archive preserves the
original endpoint values, target scope, ligand and target information and
available source metadata. Its row count exceeds the 3,228,554 BindingDB
source rows because a source row reporting more than one endpoint contributes
one archival observation for each reported endpoint. This file is provided
for provenance and reprocessing and is not the model-ready measurement table.

The file
\texttt{data/BindingDB\_multichain\_target\_chains.parquet} contains 459,989
normalized chain-level records. Each row represents one expected target chain within one multi-chain BindingDB source row. A source row reporting multiple endpoints is represented once in this archive, with the reported endpoint types retained as metadata. Available chain
sequences, protein accessions and PDB identifiers are retained. These records
are excluded from the model-ready interaction tables because BindingDB does
not provide sequences for all constituent chains.

The release also provides a column dictionary defining all fields, units, data
types and missing-value conventions. The release manifest records the input
files, checksums, software versions and preprocessing parameters. Exclusion,
attrition, construct-audit, endpoint-summary and validation reports document
the records removed or retained during processing and support independent
verification of the reported dataset statistics.

\section{Technical Validation}

\subsection{Integrity of the released dataset}

Automated checks were performed after preprocessing to verify the consistency
of the released tables. The \texttt{Measurement\_ID}, \texttt{Drug\_ID},
\texttt{Protein\_Entity\_ID} and \texttt{Interaction\_ID} values were unique
within their corresponding tables. Each interaction was also unique for its
combination of endpoint, ligand and protein entity.

Every ligand and protein entity referenced by the measurement and interaction
tables was present in the corresponding shared reference table. The combined
measurement and interaction counts agreed with the sums of the four
endpoint-specific tables. All retained ligand structures could be parsed from
their standardized representations, and all retained protein entities had a
valid BindingDB-reported sequence and a reviewed Swiss-Prot accession.

Valid classification records contained a binary target, and valid regression
records contained a finite endpoint-specific target derived from at least one
exact measurement. The preprocessing workflow was deterministic, and the
reported counts were regenerated from the fixed input files. All final
identifier, connection, schema and task-field checks passed.
\\A duplicate audit confirmed that every \texttt{Measurement\_ID} was unique and
that no combination of \texttt{BindingDB\_Source\_Row\_ID} and endpoint was
represented more than once. Identical evidence patterns within individual
source documents were retained because they may represent distinct reported
assays or source records rather than duplicates introduced during
preprocessing.
\subsection{Validation of task targets and repeated evidence}

Classification and regression targets were checked independently for each
endpoint. Interactions containing consistent threshold evidence were assigned
a valid binary label. Interactions containing evidence for both classes were
retained and marked as conflicts but were excluded from the valid
classification view. Regression targets were assigned only when at least one
exact measurement was available.

Repeated exact measurements spanning at least 100-fold were flagged as
dispersed rather than removed. This preserves experimentally reported
variability while allowing users to identify interactions that may require
additional scrutiny. The final task counts are reported in
Table~\ref{tab:task_validation}.

\begin{table}[ht]
\centering
\caption{Validation summary for the endpoint-specific classification and
regression task views.}
\label{tab:task_validation}
\scriptsize
\resizebox{\textwidth}{!}{
\begin{tabular}{lrrrrrr}
\toprule
\textbf{Endpoint} &
\textbf{Valid classification} &
\textbf{Class 1} &
\textbf{Class 0} &
\textbf{Conflicts} &
\textbf{Valid regression} &
\textbf{Dispersed regression} \\
\midrule
IC$_{50}$ & 1,500,552 & 915,116 & 585,436 & 27,532 & 1,220,359 & 12,104 \\
\(K_i\)   &   433,755 & 302,979 & 130,776 &  5,363 &   382,935 &  2,636 \\
\(K_d\)   &    83,604 &  39,912 &  43,692 &  2,176 &    56,580 &  2,112 \\
EC$_{50}$ &   187,772 & 104,110 &  83,662 &  3,616 &   154,209 &  1,396 \\
\midrule
Total     & 2,205,683 & 1,362,117 & 843,566 & 38,687 & 1,814,083 & 18,248 \\
\bottomrule
\end{tabular}
}
\end{table}
Of the 70,000 interactions without a valid binary label, 38,687 contained
conflicting class evidence and 31,313 were supported only by ambiguous
measurements.

\subsection{Validation of construct-aware aggregation}

The 9,668 retained experimental protein sequences were compared with their
corresponding canonical Swiss-Prot sequences. Most entities exactly matched
the canonical sequence, while the remaining entities included fragments,
substitution variants, insertion/deletion variants and terminal extensions
(Table~\ref{tab:sequence_categories}). These results confirm that measurements
assigned to one UniProt accession can involve different experimental
sequences.

\begin{table}[ht]
\centering
\caption{Characterization of the retained sequence-defined protein entities.}
\label{tab:sequence_categories}
\small
\begin{tabular}{lr}
\toprule
\textbf{Protein-entity category} & \textbf{Entities} \\
\midrule
Exact canonical sequence & 8,191 \\
Exact canonical fragment & 548 \\
Substitution variant & 417 \\
Fragment with variants & 328 \\
Insertion/deletion variant & 139 \\
Canonical sequence with terminal extension & 24 \\
Extension with variants & 10 \\
Complex sequence difference & 11 \\
\midrule
Total & 9,668 \\
\bottomrule
\end{tabular}
\end{table}

Construct-aware aggregation was compared with an alternative procedure in
which all protein entities sharing a UniProt accession were merged. Across the
four endpoints, construct-aware processing produced 2,275,683 interaction
records, compared with 2,212,342 under accession-only aggregation. It therefore
preserved 63,341 endpoint-specific interactions that would otherwise have been
combined.

Accession-only aggregation also increased classification conflicts from
38,687 to 42,312. A total of 4,204 endpoint-specific drug--accession groups
had classification outcomes that depended on the tested experimental sequence.
The endpoint-specific results are presented in
Table~\ref{tab:construct_validation}.

\begin{table}[ht]
\centering
\caption{Comparison of construct-aware and UniProt-accession-level
aggregation. Dependent groups are endpoint-specific drug--accession groups
whose classification outcomes vary among experimental protein sequences.}
\label{tab:construct_validation}
\small
\setlength{\tabcolsep}{5pt}
\begin{tabular}{lrrrrr}
\toprule
&
\multicolumn{2}{c}{\textbf{Interactions}} &
\multicolumn{2}{c}{\textbf{Classification conflicts}} &
\shortstack{\textbf{Construct-dependent}\\\textbf{groups}} \\
\cmidrule(lr){2-3}
\cmidrule(lr){4-5}
\textbf{Endpoint} &
\textbf{Construct} &
\textbf{Accession} &
\textbf{Construct} &
\textbf{Accession} &
\\
\midrule
IC$_{50}$ & 1,548,382 & 1,492,055 & 27,532 & 30,692 & 3,681 \\
\(K_i\)   &   445,298 &   441,518 &  5,363 &  5,598 &   244 \\
\(K_d\)   &    86,858 &    84,379 &  2,176 &  2,387 &   258 \\
EC$_{50}$ &   195,145 &   194,390 &  3,616 &  3,635 &    21 \\
\midrule
Total     & 2,275,683 & 2,212,342 & 38,687 & 42,312 & 4,204 \\
\bottomrule
\end{tabular}
\end{table}

\section{Usage Notes}

\subsection{Selecting endpoints and prediction tasks}

IC$_{50}$, \(K_i\), \(K_d\) and EC$_{50}$ should be modeled as separate
prediction targets. Users should select the endpoint-specific interaction file
corresponding to their research objective.

For classification, records should be restricted to
\texttt{classification\_valid = true}, with \texttt{y\_class} used as the
binary target. Class 1 represents evidence at or below the operational
1,000 nM threshold, whereas class 0 represents evidence above that threshold.
Interactions with conflicting or exclusively ambiguous evidence remain in the
released files but should not be used as standard binary targets.

For regression, records should be restricted to
\texttt{regression\_valid = true}. The appropriate target is
\texttt{pIC50}, \texttt{pKi}, \texttt{pKd} or \texttt{pEC50}, depending on
the selected endpoint. Interactions flagged as having dispersed exact
measurements may be retained, excluded or analyzed separately according to the
objectives of the study.

The interaction tables are recommended for model development. The measurement
tables should be used when individual observations, assay information,
inequalities, repeated evidence or source provenance are required.
Across the release, 78,449 drug--protein pairs occur under more than one
endpoint: 74,667 occur under two endpoints, 3,581 under three and 201 under all
four. Cross-endpoint experiments should therefore group identical
\texttt{Drug\_ID}--\texttt{Protein\_Entity\_ID} pairs to prevent overlap
between training and evaluation data.

\subsection{Filtering organisms and protein targets}

Organism-specific subsets should be created using \texttt{Taxon\_ID} rather
than free-text organism names. For example, human interactions can be selected
using \texttt{Taxon\_ID = 9606}. The accompanying \texttt{Organism} field is
provided for interpretation and reporting.

Protein targets can be grouped at two levels. The
\texttt{Protein\_Entity\_ID} distinguishes experimental sequences and should
be used when individual constructs, fragments or variants are treated as
separate targets. The \texttt{UniProt\_ID} groups all retained constructs
associated with the same reviewed accession.

This distinction is important when constructing evaluation partitions. A
cold-protein-entity split prevents the same experimental sequence from
appearing in more than one partition, but another construct associated with
the same UniProt accession may still occur in a different partition. A
cold-UniProt split keeps all constructs associated with one accession in the
same partition and therefore provides a stricter evaluation of generalization
to unseen proteins.

\subsection{Provenance and leakage-aware evaluation}

Evaluation partitions should be selected according to the intended
generalization question. For cold-ligand evaluation, all records sharing a
\texttt{Drug\_ID} should remain in the same partition. Cold-protein-entity and
cold-UniProt evaluations should similarly group records using
\texttt{Protein\_Entity\_ID} and \texttt{UniProt\_ID}, respectively.

Source-grouped evaluation should be constructed from the measurement tables
using \texttt{Source\_Document\_ID}. Measurements originating from the same
article, patent or other source group should remain in one partition. Because
one interaction may be supported by several source documents, source
assignment should occur at the measurement level before measurements are
aggregated within each partition.

Temporal evaluation can be constructed using \texttt{publication\_year}.
Records without a usable publication date should be handled explicitly rather
than assigned silently to a time period. Random interaction-level splitting
remains possible, but it may place related ligands, protein constructs or
measurements from the same source in different partitions.
The field \texttt{n\_measurements} counts retained BindingDB endpoint
observations and should not be interpreted as the number of statistically
independent experiments.
The release does not prescribe molecular descriptors, protein encoders or
prediction architectures. Researchers should report the selected endpoint,
task-validity filter, organism subset, identity level and splitting procedure
so that results can be interpreted and compared reproducibly.
\subsection{Limitations}

The 1,000 nM classification threshold is an operational definition rather than
a universal boundary of biological activity. Endpoint values remain dependent
on assay design and experimental conditions, and class 0 represents
threshold-negative evidence rather than confirmed absence of binding or
activity. EC$_{50}$ records do not by themselves distinguish agonist from
antagonist responses.

Construct identity depends on the sequence reported by BindingDB and therefore
cannot recover experimental differences that were not documented in the source
record. In particular, an exact match to the canonical Swiss-Prot sequence
does not prove that the complete canonical protein was used experimentally.
Conversely, sequences differing only by a reported affinity tag remain distinct
under the strict sequence-based identity used in this release. Users seeking a
more conservative evaluation should group all protein entities sharing the same
\texttt{UniProt\_ID}, particularly when constructing cold-protein splits.

Source and assay metadata are incomplete for some records, particularly
patent-derived measurements. Users should inspect the measurement-level
provenance and quality fields when constructing application-specific subsets.
\section*{Data Availability}

The Curate-DTI v1.0.0 dataset described in this article has been
deposited in Zenodo under the reserved DOI
\url{https://doi.org/10.5281/zenodo.21879265}
\cite{abouabbas2026curatedti}.
During peer review, editors and referees can access and download the
complete deposit without a Zenodo account, login, or access request
through the following link:
\textbf{\href{https://zenodo.org/records/21879265?preview=1\&token=%
eyJhbGciOiJIUzUxMiJ9.eyJpZCI6ImE1MjFmZDcxLTJiNDYtNDMxNi04NzQ0%
LWMxMDE1ZDhkYTViNCIsImRhdGEiOnt9LCJyYW5kb20iOiJkYTYwZjQ4NGUz%
MTc3ODk2ZGYyZDNmYmEyYzliOGNlZiJ9.jiPoou0d3vmULWCVzYRu0V-JFFF%
twT1mZSBqA_WqshPfJIGAQ00MkC6ox03hw2CsqRjkQiSOqvGxGvjltRD4NQ}
{Anonymous reviewer access to the Curate-DTI dataset}}.
The deposit contains endpoint-specific IC$_{50}$, $K_i$, $K_d$, and
EC$_{50}$ measurement and interaction tables; shared ligand-structure
and protein-entity tables; archival multi-chain records; a column
dictionary; preprocessing reports; validation summaries; and
file-integrity checksums. The record is currently restricted during
peer review and will be made publicly available upon publication of
this Data Descriptor. The data and accompanying documentation are
released under the Creative Commons Attribution--ShareAlike 4.0
International licence. The original BindingDB and
UniProtKB/Swiss-Prot source files are identified by release, filename,
and checksum in the accompanying manifest but are not redistributed
as unmodified source files.

\section*{Code Availability}

The preprocessing and validation software used to construct Curate-DTI
has been archived in Zenodo under the reserved DOI
\url{https://doi.org/10.5281/zenodo.22117036}
\cite{abou_abbas_2026_22117036}.
The software is also maintained in the private GitHub repository
\url{https://github.com/lina-abouabbas/Curate-DTI}, which will be made
publicly accessible upon publication of this Data Descriptor. The
archived software includes the preprocessing pipeline, validation
scripts, execution scripts, and reference software environment
required to reproduce the released dataset from the specified
BindingDB and UniProtKB/Swiss-Prot source inputs.

\section*{Competing Interests}
The authors declare no competing interests.

\section*{Acknowledgements}

Computations were performed on the Narval supercomputer, managed by Calcul
Qu\'ebec and the Digital Research Alliance of Canada. The operation of this
supercomputer is funded by the Digital Research Alliance of Canada, the
Minist\`ere de l'\'Economie, de l'Innovation et de l'\'Energie du Qu\'ebec,
and the Fonds de recherche du Qu\'ebec.

\section*{Funding}

This work was supported by the Fonds de recherche du Qu\'ebec through the
Qu\'ebec--Lebanon Scientific Research Collaboration programme
(grant 381165); the Government of Canada's New Frontiers in Research
Fund--Exploration programme (grant NFRFE-2025-00831); and the Mitacs
Globalink Research Internship programme (grant FR160654).

\bibliographystyle{naturemag}
\bibliography{references}
\end{document}